\documentclass[10pt,twocolumn,letterpaper]{article}

\usepackage[pagenumbers]{iccv} %

\definecolor{iccvblue}{rgb}{0.21,0.49,0.74}
\usepackage[pagebackref,breaklinks,colorlinks,allcolors=iccvblue]{hyperref}

\usepackage[export]{adjustbox} %
\usepackage{graphicx}
\usepackage{multirow}
\usepackage{colortbl} %
\usepackage{xcolor}   %
\usepackage[utf8]{inputenc}
\usepackage{tcolorbox}
\usepackage{tabularx}
\usepackage{ragged2e}
\usepackage{array}
\usepackage{pifont}
\usepackage{float}
\usepackage{placeins}
\usepackage{marvosym}
\title{\vspace{-1.2em}GenieBlue: Integrating both Linguistic and Multimodal Capabilities for Large Language Models on Mobile Devices\vspace{-1.2em}}

\author{
Xudong Lu$^{*1,2\dagger}$, Yinghao Chen$^{*1}$, Renshou Wu$^{*1}$, Haohao Gao$^{1}$, Xi Chen$^{1}$, Xue Yang$^{3}$, Xiangyu Zhao$^{3}$, \\
Aojun Zhou$^{2}$, Fangyuan Li$^{1}$, Yafei Wen$^{1}$, Xiaoxin Chen$^{1}$, Shuai Ren$^{1\,\ddagger\,{\textrm{\Letter}}}$, Hongsheng Li$^{2\,{\textrm{\Letter}}}$\\
\text{$^1$vivo AI Lab\quad $^2$CUHK MMLab\quad $^3$Shanghai Jiao Tong University}\\
\texttt{\{luxudong@link,hsli@ee\}.cuhk.edu.hk}\\
\texttt{shuai.ren@vivo.com}\thanks{\vspace{-1em}$^*$Equal contribution $^{\textrm{\Letter}}$Corresponding author $^\ddagger$Project lead $^\dagger$Intern at vivo.\\}
}
\vspace{-6em}
\makeatletter
\def\thanks#1{\protected@xdef\@thanks{\@thanks
        \protect\footnotetext{\hspace{-2em}#1}}}
\makeatother

\begin{document}
\maketitle
\begin{abstract}
Recent advancements in Multimodal Large Language Models (MLLMs) have enabled their deployment on mobile devices. However, challenges persist in maintaining strong language capabilities and ensuring hardware compatibility, both of which are crucial for user experience and practical deployment efficiency. In our deployment process, we observe that existing MLLMs often face performance degradation on pure language tasks, and the current NPU platforms on smartphones do not support the MoE architecture, which is commonly used to preserve pure language capabilities during multimodal training. To address these issues, we systematically analyze methods to maintain pure language capabilities during the training of MLLMs, focusing on both training data and model architecture aspects. Based on these analyses, we propose \textbf{GenieBlue}, an efficient MLLM structural design that integrates both linguistic and multimodal capabilities for LLMs on mobile devices. GenieBlue freezes the original LLM parameters during MLLM training to maintain pure language capabilities. It acquires multimodal capabilities by duplicating specific transformer blocks for full fine-tuning and integrating lightweight LoRA modules. This approach preserves language capabilities while achieving comparable multimodal performance through extensive training. Deployed on smartphone NPUs, GenieBlue demonstrates efficiency and practicality for applications on mobile devices.
\end{abstract}

\vspace*{-1em}
    
\section{Introduction}
\label{sec:intro}

Recent advancements in Large Language Models (LLMs) have significantly improved people's daily lives~\cite{jiang2024mixtral, abdin2024phi, guo2025deepseek, team2023internlm, yang2024qwen2}, particularly through multimodal models (MLLMs) that seamlessly integrate information from different sources such as text, images, and videos~\cite{openai2024gpt4o, team2024gemini, anthropic2023claude3, anil2023gemini, lu2024deepseekvl, qwen25VL, chen2024expanding}. As the scope of LLM and MLLM applications continues to expand, efficient deployment on smartphones is gaining increasing attention~\cite{xue2024powerinfer, yao2024minicpm, chu2023mobilevlm, chu2024mobilevlm, lu2024bluelm} due to their ability to enhance user privacy and support offline functionality.

In the practical process of deploying LLMs and MLLMs on smartphones, we inevitably face the storage and memory limitations inherent to these devices. Therefore, we aim to deploy a single model that can efficiently handle both pure language tasks and multimodal tasks simultaneously~\cite{qwen25VL,chen2024expanding}. Currently, various MLLMs suitable for on-device deployment have emerged, such as Qwen2.5-VL-3B~\cite{qwen25VL}, MiniCPM-V-2~\cite{yao2024minicpm}, and InternVL2.5-4B~\cite{chen2024expanding}, etc. These small models can achieve performance comparable to larger counterparts while having fewer parameters, making them ideal for on-device deployment. However, during the practical deployment of MLLMs on smartphone NPU (neural processing unit), we encounter the following issues:

\begin{table}[t]
\vspace{-1em}
\resizebox{\columnwidth}{!}{%
\begin{tabular}{llccc}
\midrule
\textbf{} & \textbf{Model} & \textbf{MATH} & \textbf{AlignBench} & \textbf{MT-Bench} \\ \midrule
Base LLM & Qwen2.5-3B & 61.74 & 6.00 & 5.81 \\ %
MLLM & InternVL2.5-4B & 55.20 & 5.18 & 4.94 \\ %
\rowcolor{gray!20}
Drop (\%) &  & 10.59 & 13.67 & 14.97 \\ \midrule
Base LLM &Qwen2.5-3B & 61.74 & 6.00 & 5.81 \\ %
MLLM & Qwen2.5-VL-3B & 58.92 & 5.38 & 4.72 \\ %
\rowcolor{gray!20}
Drop (\%) &  & 4.57 & 10.33 & 18.76 \\ \midrule
Base LLM & Qwen1.5-7B & 22.02 & 5.40 & 5.77 \\ %
MLLM & Wings-Qwen1.5-8B & 13.96 & 4.86 & 4.56 \\ %
\rowcolor{gray!20}
Drop (\%) &  & 36.60 & 10.00 & 20.97 \\ \midrule
Base LLM & BlueLM-3B & 38.94 & 5.67 & 5.42 \\  %
MLLM & GenieBlue-3B & 38.94 & 5.67 & 5.42 \\  %
\rowcolor{gray!20}
Drop (\%) &  & \textbf{0} & \textbf{0} & \textbf{0} \\ \midrule
\end{tabular}%
}
\vspace{-0.5em}
\caption{We assess the pure language capabilities of several representative MLLMs alongside their corresponding LLMs. The evaluation reveals that these MLLMs typically exhibit a performance drop exceeding 10\% across all three datasets. In contrast, our proposed GenieBlue does not sacrifice any pure language ability.}
\label{tab:nlp_down}
\vspace{-1.5em}
\end{table}

\vspace{-0.3em}
\begin{tcolorbox}[colframe=black, colback=gray!10, coltitle=black, boxrule=0.2mm, boxsep=0.2mm]
\textit{\textbf{Issue (1)} MLLMs still cannot achieve satisfactory pure language capabilities currently:}
\end{tcolorbox}
\vspace{-0.3em}
 
Current MLLMs, while excelling in multimodal tasks, still perform moderately on pure language tasks, especially in subjective language tasks, where they still exhibit significant performance gaps compared to corresponding pure language models. We here carry out a pilot study to showcase this phenomenon. We evaluate the pure language capabilities of several representative MLLMs alongside their corresponding LLMs. Both Qwen2.5-VL-3B~\cite{qwen25VL} and InternVL2.5-4B~\cite{chen2024expanding} are based on the Qwen2.5-3B~\cite{yang2024qwen2} language model. Additionally, Wings~\cite{zhang2024wings} introduces a method to train MLLMs without causing text-only forgetting. Therefore, we also assess the NLP metrics of the provided Wings-Qwen1.5 checkpoint based on Qwen1.5-7B. We select three datasets for evaluation: MATH~\cite{hendrycks2021measuring}, which consists of challenging mathematical reasoning problems; AlignBench~\cite{liu2023alignbench}, a subjective dataset for evaluating LLMs’ human alignment in Chinese; and MT-Bench~\cite{zheng2024judging}, a subjective benchmark for assessing multi-turn conversational capabilities. For the evaluation of AlignBench and MT-Bench, we leverage Google Gemini 1.5 Pro~\cite{team2024gemini} as the judge LLM. As shown in Tab.~\ref{tab:nlp_down}, these MLLMs generally suffer a drop of more than 10\% across the three datasets.

\textbf{\underline{Remark}}: For the deployment of LLMs on mobile devices, we prioritize the performance of subjective language tasks. On-device models in smartphone environments frequently engage in more nuanced, subjective tasks in daily usage, such as text refinement, call summarization, etc.

\vspace{-0.3em}
\begin{tcolorbox}[colframe=black, colback=gray!10, coltitle=black, boxrule=0.2mm, boxsep=0.2mm]
\textit{\textbf{Issue (2)} Mainstream smartphone NPU platforms currently do not support deploying MoE structures:}
\end{tcolorbox}
\vspace{-0.3em}

Currently, model structural improvements designed to integrate both multimodal and pure language capabilities typically rely on the Mixture of Experts (MoE) architecture, e.g., CogVLM~\cite{wang2023cogvlm}, Wings~\cite{zhang2024wings}. While the MoE architecture reduces the number of activated parameters during model inference, it still necessitates loading the entire original model into memory during initialization, which is not ideal for practical smartphone deployment given the limited memory available. As of now, NPU platforms of MediaTek and Qualcomm SoCs, e.g., MediaTek Dimensity 9400 and Qualcomm Snapdragon 8 Elite, do not support the deployment of MoE architectures.

Based on issue \textbf{1)}, despite extensive research into data and training methodologies for MLLMs, maintaining the pure language capability of MLLMs remains challenging. Based on issue \textbf{2)}, for end-side scenarios,  the design of model architectures must also account for the constraints imposed by deployment environments. Inspired by these two challenges, this paper systematically analyzes how to maintain pure language capabilities during the training of MLLMs from both training data and model architecture aspects, emphasizing end-side deployment considerations.

From the \underline{training data} perspective, we train LLMs using representative open-source MLLM datasets~\cite{tong2024cambrian}, consisting of 2.5M samples for pre-training and 7M for fine-tuning. Our findings reveal a significant decline in pure language capabilities. We then augment the fine-tuning dataset with an additional 2M samples of pure language data and retrain the MLLM. This modification demonstrates moderate benefits for objective NLP tasks, but it yields only minimal improvements in subjective tasks due to the currently limited volume of high-quality training data available for human preference alignment~\cite{cao2025condor,zhao2025omnialign}. From these observations, we conclude that simply increasing training data is insufficient to address the decline in pure language capabilities at the current stage. Therefore, we explore the design of \underline{model structures} considering hardware limitations of mobile NPUs. In this paper, we introduce \textbf{GenieBlue}, which integrates linguistic and multimodal capabilities for LLMs on mobile devices through efficient structural designs.

Specifically, to preserve the pure language capabilities of the original LLM, we freeze all LLM parameters during the multimodal training process. We then copy the transformer block every $n$th block for full parameter training and add LoRA~\cite{hu2021lora} modules to the remaining blocks. During inference, we adopt a non-shared base deployment approach. In the LLM inference process, we utilize the originally frozen model. For MLLM inference, we replace the original transformer blocks (every \( n \)th block) with the fully trained ones and incorporate the trained LoRA parameters.

After extensive data training, GenieBlue achieves multimodal capabilities comparable to those of fully fine-tuned MLLMs without sacrificing any pure language capabilities. We also deploy GenieBlue on the NPU of real smartphones, demonstrating its efficiency and practicality for edge computing applications on mobile devices. The contributions of our work can be summarized as follows:

\textbf{1)} We examine the deployment of MLLMs on smartphones, identifying performance degradation in text-only tasks and highlighting the limitations of current NPU platforms that do not support the deployment of MoE models.

\textbf{2)} We analyze how to maintain pure language performance during the training of MLLMs from both the training data and model structure perspectives. Then, we introduce GenieBlue, which integrates both linguistic and multimodal capabilities for LLMs on mobile devices through efficient and more hardware-friendly model structural designs.

\textbf{3)} We train GenieBlue with large amounts of multimodal datasets, achieving multimodal capabilities comparable to fully fine-tuned MLLMs without compromising any pure language abilities. We also support the deployment of GenieBlue on actual smartphone NPUs, demonstrating efficient performance in real-world mobile environments.

\begin{table*}[t]\small
\centering
\vspace{-1em}
\resizebox{\textwidth}{!}{%
\renewcommand\arraystretch{0.9}
\begin{tabular}{ll>{\RaggedRight\arraybackslash}p{13cm}}
\midrule
\textbf{Type} & \textbf{\#Samples} & \textbf{Datasets} \\ \midrule
General QA & 840k & UltraFeedback~\cite{cui2023ultrafeedback}, UltraChat~\cite{ding2023enhancing}, NoRobots~\cite{no_robots}, LIMA~\cite{zhou2023lima}, SlimOrca~\cite{SlimOrca}, WizardLM-Evol-Instruct-70K~\cite{xu2024wizardlm}, Llama-3-Magpie-Pro~\cite{xu2024magpie}, Magpie-Qwen2-Pro~\cite{xu2024magpie}, Firefly~\cite{Firefly}, Dolly~\cite{DatabricksBlog2023DollyV2}, OpenAI-Summarize-TLDR~\cite{openai2023tldr}, Know-Saraswati-CoT~\cite{2023knowrohit07} \\ \midrule
Code & 360k & Code-Feedback~\cite{zheng2024opencodeinterpreter}, Glaive-Code-Assistant~\cite{glaivecodev3}, XCoder-80K~\cite{wang2024your}, Evol-Instruct-Code~\cite{luo2023wizardcoder} \\ \midrule
Mathematics & 830k & GSM8K-Socratic~\cite{cobbe2021gsm8k}, NuminaMath-TIR~\cite{numina_math_datasets}, NuminaMath-CoT~\cite{numina_math_datasets_cot}, InfinityMATH\cite{zhang2024inifinitymath}, MathQA~\cite{amini-etal-2019-mathqa}, MetaMathQA~\cite{yu2023metamath} \\ \midrule
\end{tabular}
}
\vspace{-1em}
\caption{We expand the Cambrian-7M dataset with 2M pure text data training samples, primarily sourced from the InternVL2.5 paper~\cite{chen2024expanding}.}
\label{tab:text_data_internvl}
\vspace{-0.5em}
\end{table*}

\begin{table*}[t]\scriptsize
\renewcommand\arraystretch{0.95}
\resizebox{\textwidth}{!}{%
\begin{tabular}{llcccccccc}
\midrule
\textbf{BlueLM-3B} & \textbf{\#Samples} & \textbf{AI2D} & \textbf{ChartQA} & \textbf{DocVQA} & \textbf{OCRBench} & \textbf{RealWorldQA} & \textbf{ScienceQA} & \textbf{TextVQA} & \textbf{AVG} \\ \midrule
 & 7M & 74.81 & 68.32 & 74.60 & 55.30 & 62.35 & 67.91 & 60.06 & 66.19 \\ %
\multirow{-2}{*}{\textbf{MLLM   Tasks}} & \cellcolor[HTML]{E6E6E6}7M+2M & \cellcolor[HTML]{E6E6E6}74.03 & \cellcolor[HTML]{E6E6E6}69.36 & \cellcolor[HTML]{E6E6E6}74.63 & \cellcolor[HTML]{E6E6E6}56.70 & \cellcolor[HTML]{E6E6E6}58.04 & \cellcolor[HTML]{E6E6E6}68.24 & \cellcolor[HTML]{E6E6E6}62.34 & \cellcolor[HTML]{E6E6E6}66.19 \\ \midrule
\textbf{BlueLM-3B} & \textbf{\#Samples} & \textbf{DROP} & \textbf{GPQA} & \textbf{GSM8K} & \textbf{MATH} & \textbf{MMLU} & \textbf{AlignBench} & \textbf{MT-bench} & \textbf{AVG} \\ \midrule
 & - & 81.57 & 29.46 & 86.13 & 38.94 & 74.13 & 5.67 & 5.42 & 60.16 \\ %
 & 7M & 62.49 & 23.21 & 66.11 & 19.26 & 57.50 & 3.87 & 3.92 & 43.78 \\ %
\multirow{-3}{*}{\textbf{LLM Tasks}} & \cellcolor[HTML]{E6E6E6}7M+2M & \cellcolor[HTML]{E6E6E6}64.67 & \cellcolor[HTML]{E6E6E6}28.80 & \cellcolor[HTML]{E6E6E6}69.90 & \cellcolor[HTML]{E6E6E6}30.60 & \cellcolor[HTML]{E6E6E6}57.67 & \cellcolor[HTML]{E6E6E6}3.84 & \cellcolor[HTML]{E6E6E6}3.92 & \cellcolor[HTML]{E6E6E6}47.03 \\ \midrule\midrule
\textbf{Qwen2.5-3B} & \textbf{\#Samples} & \textbf{AI2D} & \textbf{ChartQA} & \textbf{DocVQA} & \textbf{OCRBench} & \textbf{RealWorldQA} & \textbf{ScienceQA} & \textbf{TextVQA} & \textbf{AVG} \\ \midrule
 & 7M & 77.20 & 67.36 & 68.84 & 54.70 & 61.05 & 68.19 & 57.72 & 65.01 \\ %
\multirow{-2}{*}{\textbf{MLLM   Tasks}} & \cellcolor[HTML]{E6E6E6}7M+2M & \cellcolor[HTML]{E6E6E6}76.98 & \cellcolor[HTML]{E6E6E6}68.48 & \cellcolor[HTML]{E6E6E6}64.25 & \cellcolor[HTML]{E6E6E6}56.20 & \cellcolor[HTML]{E6E6E6}62.09 & \cellcolor[HTML]{E6E6E6}69.43 & \cellcolor[HTML]{E6E6E6}55.54 & \cellcolor[HTML]{E6E6E6}64.71 \\ \midrule
\textbf{Qwen2.5-3B} & \textbf{\#Samples} & \textbf{DROP} & \textbf{GPQA} & \textbf{GSM8K} & \textbf{MATH} & \textbf{MMLU} & \textbf{AlignBench} & \textbf{MT-bench} & \textbf{AVG} \\ \midrule
 & - & 70.82 & 30.30 & 74.75 & 61.74 & 66.31 & 6.00 & 5.81 & 60.29 \\ %
 & 7M & 69.38 & 20.71 & 68.54 & 31.46 & 63.46 & 4.61 & 4.54 & 49.29 \\ %
\multirow{-3}{*}{\textbf{LLM Tasks}} & \cellcolor[HTML]{E6E6E6}7M+2M & \cellcolor[HTML]{E6E6E6}71.45 & \cellcolor[HTML]{E6E6E6}27.78 & \cellcolor[HTML]{E6E6E6}69.37 & \cellcolor[HTML]{E6E6E6}40.18 & \cellcolor[HTML]{E6E6E6}64.34 & \cellcolor[HTML]{E6E6E6}4.36 & \cellcolor[HTML]{E6E6E6}4.34 & \cellcolor[HTML]{E6E6E6}51.45 \\ \midrule
\end{tabular}%
}
\vspace{-1em}
\caption{We fully fine-tune BlueLM-V-3B from scratch (with SigLIP~\cite{zhai2023sigmoid} and BlueLM-3B~\cite{lu2024bluelm}/Qwen2.5-3B~\cite{yang2024qwen2}) using Cambrian 2.5M pre-training data and 7M fine-tuning data. We also conduct fine-tuning by adding 2M text-only data to the Cambrian-7M fine-tuning dataset. The inclusion of text-only data does not cause obvious degradation in MLLM performance and partially improves the accuracy on objective NLP tasks, but does not help with subjective NLP tasks (\#Samples denotes the number of fine-tuning data samples).}
\label{tab:full_ft_bluelm}
\vspace{-1.5em}
\end{table*}

\section{Related Works}
\label{sec:relate}

\subsection{On-device LLMs and MLLMs}

In recent years, beyond exploring scaling laws and training models with larger numbers of parameters on extensive datasets~\cite{guo2025deepseek,liu2024sphinx,chen2024expanding,qwen25VL}, a promising research direction has emerged: enabling smaller LLMs and MLLMs to achieve performance comparable to larger models~\cite{hu2024minicpm,yao2024minicpm,chu2023mobilevlm,chu2024mobilevlm,lu2024bluelm}. Small models with strong performance are more suitable for edge deployment scenarios, especially given the constraints of memory and computational resources on mobile devices like smartphones. With the advancement of this area of research, various small language models (SLMs) have been created~\cite{lu2024small}, including the Qwen series models~\cite{yang2024qwen2,yang2024qwen2technicalreport,qwen25VL,Qwen2VL}, InternLM series models~\cite{2023internlm,luo2024mono,chen2024expanding}, and the MiniCPM series models~\cite{hu2024minicpm,yao2024minicpm}. In addition to exploring methods for training SLMs with high performance, recent research has also focused on how to more effectively deploy these models on edge devices~\cite{song2023powerinfer,xue2024powerinfer}.

\subsection{Language Capability Maintenance of MLLMs}

The maintenance of original pure language capabilities during the training of MLLMs is a critical issue~\cite{wang2023cogvlm,qwen25VL,zhang2024wings}. This is particularly significant in scenarios where memory and storage are limited on edge devices, emphasizing the importance of having a model that can efficiently handle both pure language and multimodal tasks. There are now basically two types of approaches used to maintain pure language capabilities during multimodal training. The first is to increase the amount of language data during the multimodal training process~\cite{chen2024expanding,qwen25VL,lu2024deepseekvl}. However, as demonstrated by the experiments in Sec.~\ref{sec:intro}, the current approach provides limited assistance in restoring language capabilities. The second approach is to carefully design the model structure~\cite{wang2023cogvlm,luo2024mono,zhang2024wings}. Most existing methods utilize MoE architectures, which separate the ``experts'' that process text from those that handle other modal information. However, mainstream NPU platforms currently do not support the deployment of MoE structures. Recently, RL methods (e.g., DPO~\cite{rafailov2023direct}) have been utilized to align models with human preference, such as Qwen2.5-VL~\cite{qwen25VL}. However, these methods still do not fully restore the language capabilities of the model (see Tab.~\ref{tab:nlp_down} and Tab.~\ref{tab:llm_acc}). Additionally, most mainstream MLLMs still rely on pre-training and fine-tuning strategies, such as InternVL 2.5~\cite{chen2024expanding}, DeepSeek-VL2~\cite{wu2024deepseek}, Ovis~\cite{lu2024ovis}, and LLaVA-OneVision~\cite{li2024llava}. Therefore, we discuss the pre-training and fine-tuning approach in this paper.

\subsection{Benchmarks for Evaluating LLMs}

The benchmarks for assessing LLMs can now be broadly categorized into two types: objective benchmarks and subjective benchmarks. Objective benchmarks are mainly designed to directly evaluate the knowledge capabilities of LLMs, encompassing areas such as general knowledge~\cite{hendrycks2020measuring, wang2024mmlu, allenai_arc}, mathematics and science~\cite{cobbe2021training, hendrycks2021measuring, rein2023gpqa}, coding proficiency~\cite{austin2021program, chen2021evaluating}, etc. Subjective benchmarks, on the other hand, are characterized by their reliance on human judgment and interpretation~\cite{zheng2023judging,liu2023alignbench}, often requiring creativity and nuanced understanding rather than mere factual accuracy~\cite{arenahard2024,li2024crowdsourced,alpaca_eval}. For on-device deployment (e.g., on smartphones), LLMs do not necessarily need to master complex knowledge but rather require better instruction following abilities, prioritizing a stronger ability in subjective tasks.

\section{Text Capability Maintenance for MLLMs}
\label{sec:method}

In this section, we explore how to maintain the pure language capabilities during the training of MLLMs from both the training data (Sec.~\ref{sec:data_per}) and model structure perspectives (Sec.~\ref{sec:struct_per}). Based on our analyses, we propose GenieBlue (Sec.~\ref{sec:genie_design}), an efficient and hardware-friendly model structural design for MLLMs that combines both linguistic and multimodal capabilities, specifically tailored for LLMs/MLLMs on the NPUs of mobile devices.

\begin{table*}[t]\large
\vspace{-1em}
\resizebox{\textwidth}{!}{%
\begin{tabular}{lcccccccccc}
\midrule
\textbf{BlueLM-3B} & \textbf{\#Param} & \textbf{AI2D} & \textbf{ChartQA} & \textbf{DocVQA} & \textbf{OCRBench} & \textbf{RealWorldQA} & \textbf{ScienceQA} & \textbf{TextVQA} & \textbf{AVG} & \textbf{Retention (\%)} \\ \midrule
\textbf{Full-Finetune} & 3161.26M & 74.03 & 69.36 & 74.63 & 56.70 & 58.04 & 68.24 & 62.34 & 66.19 & - \\ %
\textbf{LoRA} & 458.06M & 68.23 & 61.24 & 66.17 & 48.70 & 55.56 & 68.57 & 56.97 & 60.78 & 91.82 \\ \midrule
\textbf{CogVLM-Post} & 1005.69M & 67.81 & 60.80 & 66.49 & 51.00 & 57.12 & 67.00 & 58.58 & 61.26 & 92.55 \\ %
\textbf{CogVLM-Pre} & 1005.69M & 69.04 & 64.28 & 70.23 & 51.50 & 52.29 & 67.67 & 60.42 & 62.20 & 93.98 \\ %
\rowcolor{gray!20}
\textbf{CogVLM-Skip} & 1005.69M & 70.01 & 66.36 & 71.97 & 54.60 & 56.34 & 68.91 & 59.37 & 63.94 & 96.60 \\ 
\midrule\midrule
\textbf{Qwen2.5-3B} & \textbf{\#Param} & \textbf{AI2D} & \textbf{ChartQA} & \textbf{DocVQA} & \textbf{OCRBench} & \textbf{RealWorldQA} & \textbf{ScienceQA} & \textbf{TextVQA} & \textbf{AVG} & \textbf{Retention (\%)} \\ \midrule
\textbf{Full-Finetune} & 3527.81M & 76.98 & 68.48 & 64.25 & 56.20 & 62.09 & 69.43 & 55.54 & 64.71 & - \\ %
\textbf{LoRA} & 456.84M & 65.35 & 54.32 & 55.84 & 48.10 & 55.56 & 72.72 & 58.40 & 58.61 & 90.58 \\ \midrule
\textbf{CogVLM-Post} & 1146.75M & 68.72 & 60.48 & 65.14 & 51.30 & 48.89 & 64.76 & 59.85 & 59.88 & 92.53 \\ %
\textbf{CogVLM-Pre} & 1146.75M & 68.88 & 62.12 & 67.95 & 52.30 & 53.73 & 72.87 & 57.36 & 62.17 & 96.08 \\ %
\rowcolor{gray!20}
\textbf{CogVLM-Skip} & 1146.75M & 69.30 & 65.92 & 71.10 & 54.10 & 50.59 & 69.48 & 59.62 & 62.87 & 97.16 \\  \midrule
\end{tabular}%
}
\vspace{-0.5em}
\caption{Evaluation results on MLLM benchmarks. We fine-tune all the models using the 9M dataset, comparing full fine-tuning, LoRA fine-tuning, and CogVLM fine-tuning. \textbf{Post}, \textbf{Pre}, and \textbf{Skip} means adding the visual expert module to the last quarter of the layers, the first quarter of the layers, and at every quarter interval of the layers. Apart from full fine-tuning, other methods can maintain pure language capability consistent with the original LLM during inference through the use of the non-shared base deployment strategy. CogVLM-Skip achieves the best MLLM performance retention. We also provide the trainable parameter numbers (\#Param) during MLLM training.}
\label{tab:ft_cogvlm}
\vspace{-1em}
\end{table*}

\subsection{Training Data Perspective}\label{sec:data_per}

\textbf{Approach Analysis:} To preserve pure language capabilities during the MLLM training process, the most straightforward and commonly used method is to add text-only data to the MLLM's training dataset. Currently, both InternVL2.5~\cite{chen2024expanding} and Qwen2.5-VL~\cite{qwen25VL} utilize this approach. However, this method presents some challenges. Firstly, it is difficult to collect a large amount of high-quality text-only instruction-tuning data, especially for subjective NLP tasks. Secondly, adding substantial amounts of text-only data during MLLM training will lead to longer training time.
\\[0.5em]
\hspace{-1.2em}\textbf{Quantitative Experiments:} To validate the effectiveness of this approach, we fully fine-tune an MLLM from scratch using a ViT and an LLM. Specifically, we utilize the BlueLM-V-3B architecture, which is tailored for end-side smartphone deployment, with SigLIP~\cite{zhai2023sigmoid} as the ViT and BlueLM-3B~\cite{lu2024bluelm}/Qwen2.5-3B~\cite{yang2024qwen2} as the LLM. We follow the training recipe of Cambrian-1~\cite{tong2024cambrian}, using the provided 2.5M alignment data for pre-training and the 7M data\footnote{Cambrian-7M dataset contains around 1.5M pure-text data samples.} for fine-tuning. For comparison, we add another 2M pure-text data samples to the fine-tuning dataset, primarily sourced from the InternVL2.5 paper~\cite{chen2024expanding}, as shown in Tab.~\ref{tab:text_data_internvl}. We select 7 LLM benchmarks and 7 MLLM benchmarks for evaluation. For multimodal capabilities, we choose AI2D$_\text{test}$~\cite{kembhavi2016diagram}, ChartQA$_\text{test}$~\cite{masry2022chartqa}, DocVQA$_\text{val}$~\cite{mathew2021docvqa}, OCRBench~\cite{liu2023ocrbench}, RealWorldQA~\cite{grok2024realworldqa}, ScienceQA$_\text{val}$~\cite{lu2022learn} and TextVQA$_\text{val}$~\cite{singh2019towards}. For pure language capabilities, we choose DROP$_\text{val}$~\cite{dua2019drop}, GPQA Diamond ~\cite{rein2024gpqa}, GSM8K$_\text{test}$~\cite{cobbe2021gsm8k}, MATH$_\text{test}$~\cite{hendrycks2021measuring}, MMLU$_\text{test}$~\cite{hendrycks2020measuring}, AlignBench~\cite{liu2023alignbench} and MT-Bench~\cite{zheng2024judging}. The first five LLM benchmarks assess objective language capabilities, while the last two evaluate subjective language abilities. The evaluation results are shown in Tab.~\ref{tab:full_ft_bluelm}. We come across two observations:

\vspace{-0.2em}
\begin{tcolorbox}[colframe=black, colback=gray!10, coltitle=black, boxrule=0.2mm, boxsep=0.2mm]
\textit{\textbf{Finding (1)} Adding pure-text datasets has little impact on the MLLM performance:}
\end{tcolorbox}
\vspace{-0.2em}

After adding a pure language dataset containing 2M training samples, we find that the multimodal capabilities of the trained MLLM remain virtually unchanged. This phenomenon indicates that incorporating a certain amount of pure text data during the training of an MLLM does not significantly affect its multimodal performance.

\vspace{-0.2em}
\begin{tcolorbox}[colframe=black, colback=gray!10, coltitle=black, boxrule=0.2mm, boxsep=0.2mm]
\textit{\textbf{Finding (2)} Adding pure text data leads to a moderate improvement in the performance of objective NLP tasks but does not assist with subjective tasks:}
\end{tcolorbox}
\vspace{-0.2em}

As can be seen from Tab.~\ref{tab:full_ft_bluelm}, the incorporation of multimodal data (7M) leads to a significant decline in both the objective and subjective language performance of the original LLM. To address this issue,  we refer to InternVL2.5~\cite{chen2024expanding} and integrate an additional 2M pure text samples for training. As there is still a lack of sufficient high-quality open-source training data for human alignment~\cite{cao2025condor}, the newly added pure-text data partially restores the performance for objective NLP tasks and provides almost no help for subjective NLP tasks. This indicates that maintaining the pure-language capabilities of LLMs by adding additional pure-language data currently remains a challenging endeavor.

\begin{figure}[t]
    \centering
    \vspace{-1em}
    \includegraphics[
        width=0.95\linewidth,
        margin=-1em 0 0 0,  %
        clip                 %
    ]{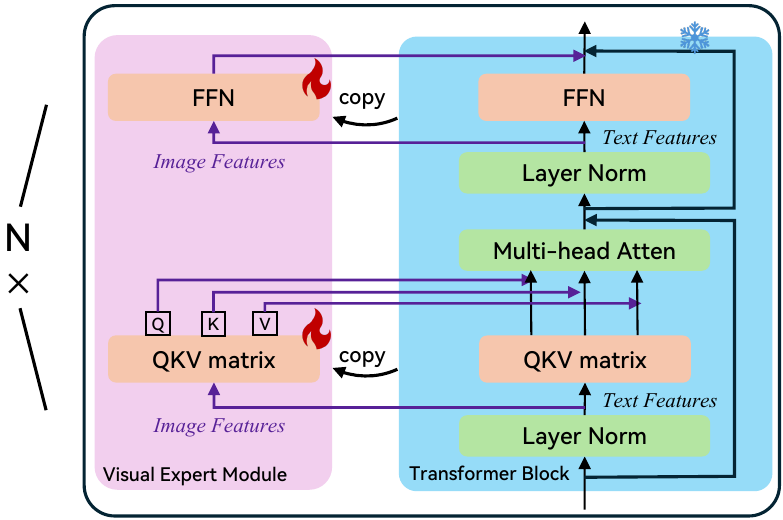}
    \caption{CogVLM~\cite{wang2023cogvlm} replicates an identical visual expert module alongside each transformer block to handle multimodal inputs.}
    \label{fig:cogvlm}
    \vspace{-1.5em}
\end{figure}

\subsection{Model Structure Perspective}\label{sec:struct_per}

\textbf{Approach Analysis:} Based on the analyses in Sec.~\ref{sec:data_per}, we conclude that maintaining NLP performance during the training of MLLMs by increasing pure text data is currently challenging. Consequently, another research direction focuses on the design of MLLM architectures, aiming to enhance NLP capabilities through architectural innovations rather than solely relying on additional pure text data. Representative works in this area include CogVLM~\cite{wang2023cogvlm} and Wings~\cite{zhang2024wings}, both of which utilize the MoE structure. 

However, during our deployment journey, we still observe that Wings~\cite{zhang2024wings} leads to a significant decline in pure language capabilities. As noted in the experiment presented in Sec.~\ref{sec:intro}, there is an average drop of over 20\% in NLP performance, which is unacceptable for our deployment purposes. Regarding CogVLM~\cite{wang2023cogvlm}, it replicates an identical visual expert module alongside each transformer block to handle multimodal inputs while keeping the original LLM frozen during training, as shown in Fig.~\ref{fig:cogvlm}. This design ensures that the performance of the original LLM remains unchanged during inference. However, this design still has two shortcomings. \textbf{1)}, during deployment, both the LLM and all corresponding visual expert modules need to be loaded into memory simultaneously, doubling the model's memory requirements. \textbf{2)}, as analyzed in Sec.~\ref{sec:intro}, current smartphone NPU platforms do not yet support the deployment of MoE models. This results in deployment issues for CogVLM on real mobile devices.
\\[0.5em]
\hspace{-1.2em}\textbf{Quantitative Experiments:} To ensure the completeness of our work, we evaluate the MLLM performance of the models after training using the CogVLM approach with both BlueLM-3B and Qwen2.5-3B LLMs. To address the memory issues that arise during deployment, we integrate a visual expert module into one-quarter of the layers. We experiment with adding visual expert modules to the last quarter of the layers, the first
quarter of the layers, and at every quarter interval of the layers~\cite{awadalla2023openflamingo}. For other transformer blocks, we add LoRA\footnote{In all experiments, we set the LoRA rank to 8.} weights to the attention modules and feed-forward modules. We compare the three CogVLM-based methods with full fine-tuning and full-LoRA training. To provide more insights, we also list the trainable parameters (including ViT and projector layer) during MLLM training. The results are shown in Tab.~\ref{tab:ft_cogvlm}.

\vspace{-0.2em}
\begin{tcolorbox}[colframe=black, colback=gray!10, coltitle=black, boxrule=0.2mm, boxsep=0.2mm]
\textit{\textbf{Finding (3)} Compared to full fine-tuning, LoRA and CogVLM methods lead to a decrease in the multimodal performance of the trained MLLM:}
\end{tcolorbox}
\vspace{-0.2em}

Due to limitations in the number of trainable parameters, both LoRA and CogVLM methods fall short of the multimodal performance achieved by full fine-tuning. Nevertheless, they typically reach over 90\% of the performance seen with full fine-tuning. Besides, CogVLM outperforms LoRA in MLLM performance. It is important to note that full fine-tuning has a significant negative impact on the performance of pure-text tasks (Tab.~\ref{tab:full_ft_bluelm}), while LoRA and CogVLM do not influence the pure language performance through the use of the non-shared base deployment strategy (Sec.~\ref{sec:genie_design}).

\vspace{-0.2em}
\begin{tcolorbox}[colframe=black, colback=gray!10, coltitle=black, boxrule=0.2mm, boxsep=0.2mm]
\textit{\textbf{Finding (4)} For CogVLM, the addition of visual expert modules at every quarter interval of the layers results in the best MLLM performance:}
\end{tcolorbox}
\vspace{-0.2em}

Incorporating visual experts at every quarter interval of the layers results in over 96\% accuracy retention for the MLLM compared to full fine-tuning. Since CogVLM's training approach does not affect pure-text performance, we have decided to design GenieBlue based on this method.

\begin{figure*}[t]
    \centering
    \vspace{-1.2em}
    \includegraphics[
        width=\linewidth,
        margin=0 0 0 0,  %
        clip                 %
    ]{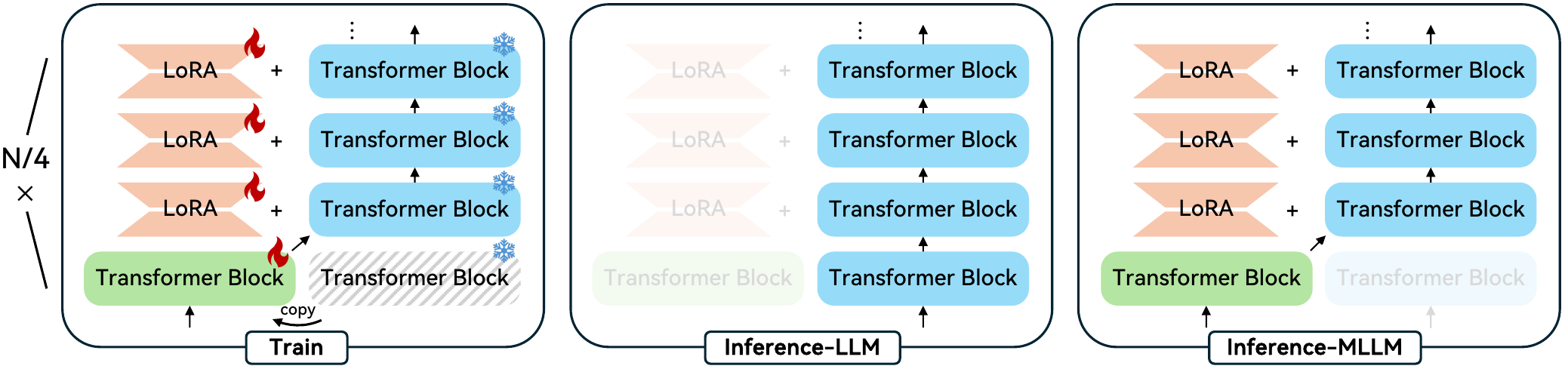}
    \caption{Overview of GenieBlue. We replicate the transformer blocks at every quarter interval of the layers in the LLM and incorporate LoRA modules into the other transformer blocks. During multimodal training, we freeze the original LLM while fully training the replicated transformer blocks and the added LoRA parameters. For pure-text inference, we utilize the original LLM. For multimodal inference, we replace the original blocks with the trained transformer blocks at every quarter interval and add LoRA to the remaining transformer blocks. This \underline{non-shared base} approach avoids the MoE structure while decoupling the inference processes of the LLM and MLLM.}
    \label{fig:overview_blue}
    \vspace{-1em}
\end{figure*}

\begin{table*}[t]\large
\resizebox{\textwidth}{!}{%
\begin{tabular}{lcccccccccc}
\midrule
\textbf{BlueLM-3B} & \textbf{\#Param} & \textbf{AI2D} & \textbf{ChartQA} & \textbf{DocVQA} & \textbf{OCRBench} & \textbf{RealWorldQA} & \textbf{ScienceQA} & \textbf{TextVQA} & \textbf{AVG} & \textbf{Retention (\%)} \\ \midrule
\textbf{Full-Finetune} & 3161.26M & 74.03 & 69.36 & 74.63 & 56.70 & 58.04 & 68.24 & 62.34 & 66.19 & - \\ %
\textbf{CogVLM-Skip} & 1005.69M & 70.01 & 66.36 & 71.97 & 54.60 & 56.34 & 68.91 & 59.37 & 63.94 & 96.60 \\ \midrule
\textbf{GenieBlue-Post} & 1005.73M & 68.49 & 61.68 & 67.78 & 49.80 & 55.42 & 69.96 & 61.59 & 62.10 & 93.82 \\ %
\textbf{GenieBlue-Pre} & 1005.73M & 72.90 & 66.20 & 71.11 & 46.50 & 58.30 & 73.20 & 60.03 & 64.03 & 96.74 \\ %
\rowcolor{gray!20}
\textbf{GenieBlue-Skip} & 1005.73M & 73.67 & 69.32 & 74.26 & 55.30 & 57.39 & 68.34 & 60.37 & 65.52 & 98.99 \\ \midrule\midrule
\textbf{Qwen2.5-3B} & \textbf{\#Param} & \textbf{AI2D} & \textbf{ChartQA} & \textbf{DocVQA} & \textbf{OCRBench} & \textbf{RealWorldQA} & \textbf{ScienceQA} & \textbf{TextVQA} & \textbf{AVG} & \textbf{Retention (\%)} \\ \midrule
\textbf{Full-Finetune} & 3527.81M & 76.98 & 68.48 & 64.25 & 56.20 & 62.09 & 69.43 & 55.54 & 64.71 & - \\ %
\textbf{CogVLM-Skip} & 1146.75M & 69.30 & 65.92 & 71.10 & 54.10 & 50.59 & 69.48 & 59.62 & 62.87 & 97.16 \\ \midrule
\textbf{GenieBlue-Post} & 1146.79M & 67.29 & 59.80 & 60.70 & 49.30 & 56.47 & 75.35 & 59.88 & 61.26 & 94.66 \\ %
\textbf{GenieBlue-Pre} & 1146.79M & 69.01 & 58.44 & 56.65 & 43.90 & 58.04 & 75.01 & 62.19 & 60.46 & 93.44 \\ %
\rowcolor{gray!20}
\textbf{GenieBlue-Skip} & 1146.79M & 72.99 & 63.04 & 62.74 & 53.90 & 57.39 & 71.05 & 61.68 & 63.26 & 97.76 \\ \midrule
\end{tabular}%
}
\vspace{-0.5em}
\caption{Evaluation results on MLLM benchmarks after training with the 9M fine-tuning dataset. Similar to the experiment setting of CogVLM, we replicate transformer blocks at the last, first, and every interval quarter of layers. Results show that GenieBlue-Skip demonstrates the best MLLM performance, yielding over 97\% retention in MLLM performance compared to full fine-tuning.}
\vspace{-1em}
\label{sec:genie_9m}
\end{table*}

\begin{table*}[t]\small
\vspace{-1.2em}
\renewcommand\arraystretch{0.9}
\resizebox{\textwidth}{!}{%
\begin{tabular}{lcccccccccc}
\midrule
\textbf{BlueLM-3B} & \textbf{Shared Base} & \textbf{DROP} & \textbf{GPQA} & \textbf{GSM8K} & \textbf{MATH} & \textbf{MMLU} & \textbf{AlignBench} & \textbf{MT-bench} & \textbf{AVG} & \textbf{Retention (\%)} \\ \midrule
\textbf{BlueLM-3B} & - & 81.57 & 29.46 & 86.13 & 38.94 & 74.13 & 5.67 & 5.42 & 60.16 & - \\ \midrule
\textbf{Full-Finetune} & - & 64.67 & 28.80 & 69.90 & 30.60 & 57.67 & 3.84 & 3.92 & 47.03 & 78.18 \\ %
\textbf{LoRA} & \checkmark & 79.71 & 29.02 & 84.46 & 39.08 & 69.76 & 4.62 & 4.61 & 56.33 & 93.63 \\ \midrule
\textbf{GenieBlue-Post} & \checkmark & 78.64 & 28.13 & 85.37 & 37.08 & 70.77 & 4.51 & 4.65 & 55.94 & 92.98 \\ %
\textbf{GenieBlue-Pre} & \checkmark & 76.95 & 29.24 & 74.98 & 35.66 & 65.26 & 4.61 & 4.71 & 53.61 & 89.12 \\ %
\textbf{GenieBlue-Skip} & \checkmark & 75.36 & 29.02 & 76.27 & 38.16 & 67.78 & 4.66 & 4.76 & 54.40 & 90.42 \\ \midrule
\rowcolor{gray!20}
\textbf{GenieBlue} & \ding{55} & 81.57 & 29.46 & 86.13 & 38.94 & 74.13 & 5.67 & 5.42 & 60.16 & 100.00 \\ \midrule
\midrule
\textbf{Qwen2.5-3B} & \textbf{Shared Base} & \textbf{DROP} & \textbf{GPQA} & \textbf{GSM8K} & \textbf{MATH} & \textbf{MMLU} & \textbf{AlignBench} & \textbf{MT-bench} & \textbf{AVG} & \textbf{Retention (\%)} \\ \midrule
\textbf{Qwen2.5-3B} & - & 70.82 & 30.30 & 74.75 & 61.74 & 66.31 & 6.00 & 5.81 & 60.29 & - \\ \midrule
\textbf{Full-Finetune} & - & 71.45 & 27.78 & 69.37 & 40.18 & 64.34 & 4.36 & 4.34 & 51.45 & 85.33 \\ %
\textbf{LoRA} & \checkmark & 53.94 & 23.74 & 70.96 & 43.94 & 66.00 & 4.17 & 4.36 & 49.13 & 81.48 \\ \midrule
\textbf{GenieBlue-Post} & \checkmark & 60.97 & 20.20 & 72.48 & 43.10 & 64.84 & 4.31 & 4.90 & 50.53 & 83.81 \\ %
\textbf{GenieBlue-Pre} & \checkmark & 61.65 & 27.27 & 72.25 & 42.94 & 66.99 & 4.45 & 4.69 & 51.79 & 85.90 \\ %
\textbf{GenieBlue-Skip} & \checkmark & 67.98 & 28.28 & 69.90 & 42.64 & 65.97 & 4.62 & 4.70 & 52.57 & 87.19 \\ \midrule
\rowcolor{gray!20}
\textbf{GenieBlue} & \ding{55} & 70.82 & 30.30 & 74.75 & 61.74 & 66.31 & 6.00 & 5.81 & 60.29 & 100.00 \\ \midrule
\end{tabular}%
}
\vspace{-0.5em}
\caption{Comparison of pure language capabilities using the shared base versus non-shared base deployment strategies, trained with 9M fine-tuning data. The non-shared base approach can maintain the pure text capabilities of the original LLM. In the shared-base strategy, training with BlueLM-3B indicates that the fewer trainable parameters involved in multimodal training, the better the retention of pure text capabilities. However, the LoRA-trained MLLM based on Qwen2.5-3B achieves the worst pure-text performance.}
\label{tab:share_nlp}
\vspace{-1em}
\end{table*}

\subsection{GenieBlue}\label{sec:genie_design}

Based on the analyses from both data (Sec.~\ref{sec:data_per}) and structural (Sec.~\ref{sec:struct_per}) perspectives, we propose to integrate linguistic and multimodal capabilities into the training of MLLMs through structural design. In this subsection, we provide a detailed illustration of the GenieBlue structure.
\\[0.5em]
\textbf{Approach Analysis:} We modify from the CogVLM~\cite{wang2023cogvlm} structure, particularly paying attention to the limitations of NPUs on the MoE architecture. The main idea behind CogVLM is to separate the processing of text tokens and multimodal tokens. It employs an MoE architecture where different experts handle text and visual tokens. In contrast, our design principle focuses on bypassing the MoE structure by \underline{selecting separate model weights} for LLM/MLLM deployment, thereby maintaining the original LLM architecture unchanged during the multimodal inference process.

The framework of GenieBlue is shown in Fig.~\ref{fig:overview_blue}. To save model storage on smartphones, we replicate the transformer blocks at every quarter interval throughout the layers of the LLM while integrating LoRA modules into the remaining transformer blocks. During multimodal training, we freeze the original LLM, allowing ViT, the replicated transformer blocks, and the added LoRA parameters to be fully trained.

For pure-text inference, we utilize the original, unmodified LLM to perform all calculations. In contrast, for multimodal inference, we replace the original blocks with the trained transformer blocks at every quarter interval and incorporate LoRA into the remaining transformer blocks. This \underline{non-shared base strategy} effectively avoids the MoE structure and decouples the inference processes of the LLM and MLLM. During actual NPU deployment, we only need to replace the weights and adapt the LoRA module. This makes deployment simple and efficient.
\\[0.5em]
\hspace{-1.2em}\textbf{Quantitative Experiments:} We compare our proposed GenieBlue against full fine-tuning and the CogVLM methods with both BlueLM-3B and Qwen2.5-3B LLMs, using the 2.5M pre-training data and 9M fine-tuning data. For a fair comparison with CogVLM, we replicate transformer blocks at the last (Post), first (Pre), and every interval (Skip) quarter of layers. The results are shown in Tab.~\ref{sec:genie_9m}.

\vspace{-0.2em}
\begin{tcolorbox}[colframe=black, colback=gray!10, coltitle=black, boxrule=0.2mm, boxsep=0.2mm]
\textit{\textbf{Finding (5)} For GenieBlue structure, GenieBlue-Skip achieves the best multimodal performance, GenieBlue-Skip also outperforms CogVLM-Skip:}
\end{tcolorbox}
\vspace{-0.2em}

Similar to the results of CogVLM, replicating transformer blocks at every interval quarter of layers achieves better multimodal performance. Besides, we find that GenieBlue-Skip outperforms CogVLM-Skip. This could possibly be attributed to CogVLM's approach of incorporating visual expert modules. In CogVLM's design, text features and image features are rigidly separated and processed separately for QKV and FFN calculations. Although CogVLM considers the fusion of text and image features during multi-head attention, this fusion is not as effective as completely sharing weights, which limits better integration throughout the entire MLLM inference process.
\\[0.5em]
\hspace{-1.2em}\textbf{Non-shared Base Deployment Strategy:} By splitting the LLM and MLLM inference process, deploying GenieBlue with the non-shared base strategy (as shown in Fig.~\ref{fig:overview_blue}) can maintain the pure language capabilities of the original LLM. To validate the importance of this approach, we evaluate GenieBlue's performance on LLM benchmarks, comparing the shared and non-shared base deployment strategies. The shared base deployment strategy refers to unifying the inference processes of LLM and MLLM into the single deployment mode depicted on the right of Fig.~\ref{fig:overview_blue}. Specifically, during the inference of pure language tasks, we also leverage the fully trained transformer blocks and incorporate the LoRA module.  Additionally, we provide the NLP performances of BlueLM-3B/Qwen2.5-3B, the fully fine-tuned models, and the models trained entirely with LoRA. The results are shown in Tab.~\ref{tab:share_nlp}. 
\vspace{-0.2em}
\begin{tcolorbox}[colframe=black, colback=gray!10, coltitle=black, boxrule=0.2mm, boxsep=0.2mm]
\textit{\textbf{Finding (6)} Deploying with the non-shared base strategy results in significantly better pure-text capabilities compared to the shared base strategy:}
\end{tcolorbox}
\vspace{-0.2em}

Undoubtedly, using the shared base deployment strategy leads to a loss of pure language capabilities, demonstrating the importance of the non-shared base deployment method. Another interesting finding is that, intuitively, with the same MLLM training data, having fewer trainable parameters results in less loss of the model's pure language performance. Training with BlueLM-3B aligns with this intuition. However, the LoRA-trained MLLM based on Qwen2.5-3B achieves the worst pure-text performance. A plausible explanation for this phenomenon lies in the inherent mechanism of LoRA, which imposes low-rank matrices onto original weights rather than directly training the base parameters. The limited number of adapter parameters may hinder effective integration with the pre-existing model parameters, resulting in suboptimal parameter fusion and consequently injuring the LLM performance.

\section{Training and Deployment Recipe}

After analyzing from both training data and model structure perspectives in Sec.~\ref{sec:method}, we determine the model structure (GenieBlue-Skip) and deployment approach (non-shared base deployment strategy). In this section, we introduce the detailed training (Sec.~\ref{sec:train_rec}) and deployment details (Sec.~\ref{sec:deploy_rec}) of the final GenieBlue model. 

\subsection{Training Recipe}\label{sec:train_rec}

We employ the GenieBlue-Skip structure and strictly adhere to the training recipe and training data of BlueLM-V-3B~\cite{lu2024bluelm}. Specifically, our training process consists of two stages. In the first stage, we pre-train the MLP projection layer while keeping the ViT and LLM frozen, using the 2.5M pre-training data. In the second stage, we fine-tune the GenieBlue-Skip model (ViT, projector, replicated transformer blocks, and the added LoRA parameters) with 645M fine-tuning data~\cite{lu2024bluelm} while keeping the original LLM frozen. We use SigLIP as the ViT and BlueLM-3B as the LLM. During training, we set the LoRA rank to 8.

\subsection{Deployment Recipe}\label{sec:deploy_rec}

We deploy GenieBlue on the NPU of the iQOO 13 smartphone, which is equipped with the Qualcomm Snapdragon 8 Elite (Gen 4) SoC. We leverage the Qualcomm QNN SDK\footnote{\url{https://www.qualcomm.com/developer/software/neural-processing-sdk-for-ai}} for model deployment. For the ViT and projector layer, we employ W8A16 quantization. For the LLM, we adopt W4A16 quantization. Regarding the added LoRA parameters, we utilize a W8A16 quantization scheme. Currently, we support the single-patch ViT inference. It is important to note that the Snapdragon 8 Elite's NPU platform does not support the deployment of MoE structures.

\section{Performance of GenieBlue}
\label{sec:experiment}

\begin{table*}[t]\Large
\vspace{-1.2em}
\renewcommand\arraystretch{1}
\resizebox{\textwidth}{!}{%
\begin{tabular}{lcccccccccc}
\midrule
\textbf{Model} & \textbf{\#Params} & \textbf{AVG} & \textbf{MMBench} & \textbf{MMStar} & \textbf{MMMU} & \textbf{MathVista} & \textbf{HallusionBench} & \textbf{AI2D} & \textbf{OCRBench} & \textbf{MMVet} \\ \midrule
\textbf{BlueLM-V-3B}~\cite{lu2024bluelm} & 3.2B & 66.1 & 82.7 & 62.3 & 45.1 & 60.9 & 48.0 & 85.3 & 82.9 & 61.8 \\ \midrule
\textbf{Ovis2-2B}~\cite{lu2024ovis} & 2.46B & 65.2 & 76.9 & 56.7 & 45.6 & 64.1 & 50.2 & 82.7 & 87.3 & 58.3 \\ %
\textbf{Qwen2.5-VL-3B}~\cite{qwen25VL} & 3.75B & 64.5 & 76.8 & 56.3 & 51.2 & 61.2 & 46.6 & 81.4 & 82.8 & 60.0 \\ %
\textbf{SAIL-VL-2B}~\cite{dong2025scalable} & 2.1B & 61.0 & 73.7 & 56.5 & 44.1 & 62.8 & 45.9 & 77.4 & 83.1 & 44.2 \\ %
\textbf{InternVL2.5-2B-MPO}~\cite{wang2024mpo} & 2B & 60.9 & 70.7 & 54.9 & 44.6 & 53.4 & 40.7 & 75.1 & 83.8 & 64.2 \\ %
\rowcolor{gray!20}
\textbf{GenieBlue} & 3.2(+0.55)B & 64.2 & 78.2 & 59.4 & 47.6 & 58.0 & 46.3 & 83.1 & 82.9 & 58.1 \\ \midrule
\textbf{InternVL2-8B}~\cite{chen2024far} & 8B & 64.1 & 79.4 & 61.5 & 51.2 & 58.3 & 45.0 & 83.6 & 79.4 & 54.3 \\ \midrule
\end{tabular}%
}
\vspace{-0.5em}
\caption{Performance on MLLM benchmarks under the same evaluation settings as OpenCompass benchmark ($\leq$ 4B, with InternVL2-8B for reference). GenieBlue retains over 97\% accuracy of BlueLM-V-3B while outperforming InternVL2-8B on average. $^\dagger$The total number of parameters in the replicated transformer blocks and LoRA modules is 0.55B.}
\label{tab:opencompass}
\vspace{-0.8em}
\end{table*}

\begin{table*}[t]\scriptsize
\resizebox{\textwidth}{!}{%
\begin{tabular}{lcccccccccc}
\midrule
 & \textbf{\#Params} & \textbf{DROP} & \textbf{GPQA} & \textbf{GSM8K} & \textbf{MATH} & \textbf{MMLU} & \textbf{AlignBench} & \textbf{MT-bench} & \textbf{AVG} & \textbf{Retention (\%)} \\ \midrule
\textbf{BlueLM-3B} & 2.7B & 81.57 & 29.46 & 86.13 & 38.94 & 74.13 & 5.67 & 5.42 & 60.16 & - \\ %
\rowcolor{gray!20}
\textbf{GenieBlue} & 3.2(+0.55)B & 81.57 & 29.46 & 86.13 & 38.94 & 74.13 & 5.67 & 5.42 & 60.16 & 100.00 \\ \midrule
\textbf{Qwen2.5-3B} & 3.1B & 70.82 & 30.30 & 74.75 & 61.74 & 66.31 & 6.00 & 5.81 & 60.29 & - \\ %
\rowcolor{gray!20}
\textbf{Qwen2.5VL-3B} & 3.75B & 72.72 & 24.24 & 70.43 & 58.92 & 65.07 & 5.38 & 4.72 & 56.05 & 92.98 \\ \midrule
\end{tabular}%
}
\vspace{-1em}
\caption{Evaluation results on representative LLM benchmarks, including both objective and subjective benchmarks. GenieBlue retains 100\% performance of the original LLM, whereas Qwen2.5VL-3B exhibits some degradation.}
\label{tab:llm_acc}
\vspace{-0.5em}
\end{table*}

\begin{table*}[h!t]%
\vspace{-0.5em}
\resizebox{\textwidth}{!}{%
\begin{tabular}{lcccccccc}
\midrule
\textbf{Model} & \textbf{Context (token)} & \textbf{Load Time (s)} & \textbf{ViT Time (s)} & \textbf{Input Speed (token/s)} & \textbf{Output Speed (token/s)} & \textbf{Storage (GB)} & \textbf{Memory (GB)} \\ \midrule
\textbf{BlueLM-V-3B} & 2048 & 0.51 & 0.4 & 1515.15 & 33.00 & 1.77 & 1.73 \\ %
\rowcolor{gray!20}
\textbf{GenieBlue} & 2048 & 0.80 & 0.4 & 1666.67 & 31.00 & 1.92 & 2.10 \\ \midrule
\end{tabular}%
}
\vspace{-0.5em}
\caption{Deployment efficiency comparison between GenieBlue and BlueLM-V-3B on Qualcomm 8 Elite SoC in peak performance mode. GenieBlue results in a longer model loading time, slightly higher storage and memory usage, and a marginally slower token output speed.}
\label{tab:deploy_data}
\vspace{-1.4em}
\end{table*}

Through extensive data training and NPU deployment, in this section, we evaluate the MLLM (Sec.~\ref{sec:mllm_res}) and LLM (Sec.~\ref{sec:llm_res}) capabilities of GenieBlue, as well as its deployment efficiency on smartphone NPUs (Sec.~\ref{sec:deploy_res}).

\subsection{MLLM Performance}\label{sec:mllm_res}

After extensive data training, we evaluate our model using representative MLLM benchmarks, including MMbench~\cite{liu2025mmbench}, MMStar~\cite{chen2024we}, MMMU~\cite{yue2024mmmu}, MathVista~\cite{lu2023mathvista}, HallusionBench~\cite{guan2023hallusionbench}, AI2D~\cite{kembhavi2016diagram}, OCRBench~\cite{liu2023ocrbench}, and MM-Vet~\cite{yu2023mm}, which are integrated into the OpenCompass benchmark suite~\cite{2023opencompass}. We compare GenieBlue with other MLLMs that have fewer than 4B parameters, and the results are presented in Tab.~\ref{tab:opencompass}. GenieBlue achieves MLLM accuracy slightly lower than Qwen2.5-VL-3B while retaining 97\% performance of BlueLM-V-3B. Besides, GenieBlue slightly outperforms InternVL2-8B on average.

\subsection{LLM Performance}\label{sec:llm_res}

The most significant feature of GenieBlue is that it does not lose LLM performance when deployed using the non-shared base deployment strategy. Here, we evaluate its LLM performance on representative benchmarks. For comparison, we select Qwen2.5VL-3B, which claims to maintain LLM performance without degradation from MLLM training by incorporating pure-text data. As demonstrated in Tab.~\ref{tab:llm_acc}, GenieBlue achieves no loss in LLM performance, while Qwen2.5VL-3B exhibits some performance degradation, especially in subjective tasks. This indicates that exploring model structure design is more effective for maintaining pure-text capabilities than simply increasing the amount of pure-text data currently.

\subsection{Deployment Efficiency}\label{sec:deploy_res}

We deploy GenieBlue with the non-shared base strategy on Qualcomm Snapdragon 8 Elite (Gen 4) SoC. Different from~\cite{lu2024bluelm}, we now support the 1-patch ViT inference. We here provide the MLLM deployment statistics in Tab.~\ref{tab:deploy_data}, comparing BlueLM-V-3B and GenieBlue. With the inclusion of additional LoRA parameters, GenieBlue incurs longer model loading times, slightly larger storage and memory requirements, and a marginally slower token output speed. However, a token output speed of 30 token/s is fully sufficient for daily use on mobile devices.

\section{Conclusion}
\label{sec:conclusion}

In this paper, we approach the challenge of maintaining pure language capabilities from a practical deployment perspective on mobile devices (smartphones), analyzing both training data and model structure to identify effective strategies. Based on the analyses, we propose GenieBlue, an efficient and hardware-friendly MLLM design that integrates linguistic and multimodal capabilities for mobile LLMs. By freezing the original LLM parameters during training and acquiring multimodal capabilities through duplicated transformer blocks and lightweight LoRA modules, GenieBlue maintains language performance while achieving competitive multimodal results. Deployed on smartphone NPUs, GenieBlue demonstrates its practicality and efficiency, making it a promising solution for edge computing applications on mobile devices. We hope that our work will provide valuable insights for future research in this field.

\clearpage
{
    \small
    \bibliographystyle{ieeenat_fullname}
    \bibliography{main}
}
\clearpage
\appendix
\onecolumn

\begin{minipage}{\textwidth}
\begin{figure}[H]
\vspace{-1em}
    \centering
    \includegraphics[
        width=\textwidth,
        margin=-3.5em 0 0 0,  %
        clip                 %
    ]{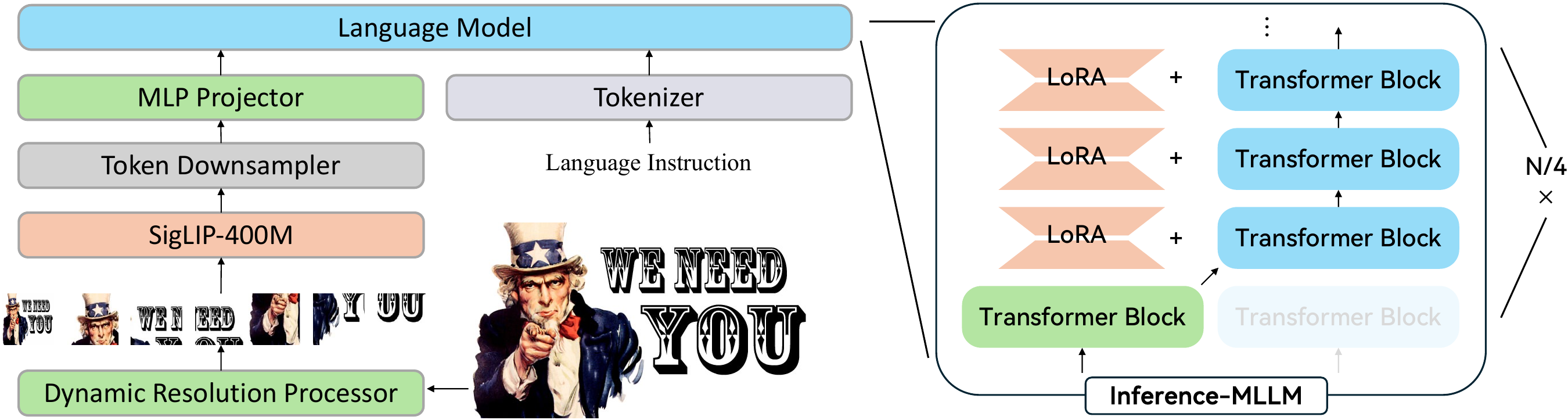}
    \caption{Structure detail of GenieBlue during the MLLM inference process.}
    \label{fig:detail_struct}
    \vspace{1em}
\end{figure}
\end{minipage}

\section{Structure Details of GenieBlue}

We here provide the detailed structure of GenieBlue during the MLLM inference process based on the BlueLM-V-3B~\cite{lu2024bluelm} architecture (Fig.~\ref{fig:detail_struct}). BlueLM-V-3B is modified from the classical LLaVA approach~\cite{liu2023llava}, incorporating a redesigned dynamic resolution processor and a token downsampler~\cite{lin2023vila} to optimize for better on-device deployment. GenieBlue further focuses on the structural design of the transformer blocks within the language model.

\section{Training Data Composition}

We here provide the data composition of Cambrian-7M~\cite{tong2024cambrian}. It has already included approximately 1.5M pure text training samples. The data composition of the 645M fine-tuning data for GenieBlue can be found in~\cite{lu2024bluelm}.

\begin{table}[h]\Huge
\centering
\resizebox{0.5\columnwidth}{!}{%
\begin{tabular}{lccccccc}
\midrule
\textbf{Type} & \textbf{OCR} & \textbf{General} & \textbf{Language} & \textbf{Counting} & \textbf{Code} & \textbf{Math} & \textbf{Science} \\ \midrule
\textbf{Ratio (\%)} & 27.22 & 34.52 & 21.00 & 8.71 & 0.87 & 7.20 & 0.88 \\ \midrule
\end{tabular}%
}
\caption{Data composition of the Cambrian-7M~\cite{tong2024cambrian} fine-tuning dataset (with approximately 1.5M pure-text data).}
\end{table}

\section{More Discussions}

GenieBlue is a plug-and-play training approach that efficiently decouples multimodal training parameters from the original language model. This design allows GenieBlue to achieve good multimodal performance without compromising the language model's performance.  In addition, this structural design requires minimal hardware-side adaptation and reduces the engineering difficulty during practical end-side deployment, making it a relatively reasonable approach at the current stage. In the future, we will validate the feasibility of GenieBlue on a wider range of SoC platforms.

\clearpage
\section{Hyper Parameters}

Here, we provide the hyper-parameters used in the pre-training and fine-tuning stage of the final GenieBlue model. We use the same 2.5M pre-training data and 645M fine-tuning data as in BlueLM-V-3B~\cite{lu2024bluelm}.

\subsection{Pre-training Stage}

\begin{table}[ht]\small
\vspace{-1em}
\centering
\resizebox{0.5\columnwidth}{!}{%
\begin{tabular}{lc}
\midrule
\textbf{Configuration} & \textbf{Stage   1} \\ \midrule
\textbf{LLM Sequence Length} & 4096 \\ \midrule
\textbf{Dynamic Resolution} & None (384$\times$384) \\ \midrule
\textbf{Optimizer} & AdamW \\ \midrule
\textbf{Optimizer Hyperparams} & $\beta_1=0.9$, $\beta_2=0.98$,   $\epsilon=10^{-6}$ \\ \midrule
\textbf{Peak LR} & $10^{-3}$ \\ \midrule
\textbf{LR Schedule} & Cosine Decay \\ \midrule
\textbf{Weight Decay} & 0.05 \\ \midrule
\textbf{Training Steps} & 3.434k \\ \midrule
\textbf{Warm-up Steps} & 34 \\ \midrule
\textbf{Global Batch Size} & 720 \\ \midrule
\textbf{Gradient Accumulation} & 1 \\ \midrule
\textbf{Numerical Precision} & \texttt{bfloat16} \\ \midrule
\end{tabular}%
}
\caption{Hyper-parameters for the pre-training stage (stage 1) of GenieBlue with 2.5M training samples.}
\label{tab:stage1_param}
\end{table}

\subsection{Fine-tuning Stage}

In the process of fine-tuning, to enhance the speed of training, we concatenate training samples to achieve a sequence length of 4096.

\begin{table}[ht]\small
\centering
\resizebox{0.5\columnwidth}{!}{%
\begin{tabular}{lc}
\midrule
\textbf{Configuration} & \textbf{Stage   2} \\ \midrule
\textbf{LLM   Sequence Length} & 4096 \\ \midrule
\textbf{Dynamic Resolution} & Up to 16 patches (1536$\times$1536) \\ \midrule
\textbf{Optimize}r & AdamW \\ \midrule
\textbf{Optimizer Hyperparams} & $\beta_1$ = 0.9, $\beta_2$ = 0.98, $\epsilon$ = 10$^{-6}$ \\ \midrule
\textbf{Peak LR} & 10$^{-4}$ \\ \midrule
\textbf{LR Schedule} & Cosine Decay \\ \midrule
\textbf{Weight Decay} & 0.05 \\ \midrule
\textbf{ViT Layer-wise LR Decay} & 0.9 \\ \midrule
\textbf{Training Steps} & 53k \\ \midrule
\textbf{Warm-up Steps} & 530 \\ \midrule
\textbf{Global Batch Size} & 6800 \\ \midrule
\textbf{Gradient Accumulation} & 10 \\ \midrule
\textbf{Numerical Precision} & \texttt{bfloat16} \\ \midrule
\end{tabular}%
}
\caption{Hyper-parameters for the fine-tuning stage (stage 2) of GenieBlue with 645M training samples.}
\vspace{-1em}
\end{table}

\end{document}